\documentclass[10pt,twocolumn,letterpaper]{article}

\usepackage[pagenumbers]{wacv} % To force page numbers, e.g. for an arXiv version

\usepackage{graphicx}
\usepackage{amsmath}
\usepackage{amssymb}
\usepackage{booktabs}
\usepackage{enumitem}
\usepackage[accsupp]{axessibility}  % Improves PDF readability for those with disabilities.

\usepackage[pagebackref,breaklinks,colorlinks]{hyperref}

\usepackage[capitalize]{cleveref}
\crefname{section}{Sec.}{Secs.}
\Crefname{section}{Section}{Sections}
\Crefname{table}{Table}{Tables}
\crefname{table}{Tab.}{Tabs.}

\def\wacvPaperID{1510} % *** Enter the WACV Paper ID here
\def\confName{WACV}
\def\confYear{2024}

\begin{document}

\title{Text-to-image Editing by Image Information Removal}

\author{Zhongping Zhang$^1$ \qquad Jian Zheng$^2$ \qquad Jacob Zhiyuan Fang$^2$ \qquad Bryan A. Plummer$^1$ \\
$^1$Boston University \qquad $^2$Amazon Alexa AI\\
{\small $^1$\texttt{\{zpzhang,bplum\}@bu.edu} \qquad $^2$\texttt{\{nzhengji, zyfang\}@amazon.com}}
}
\maketitle

%%%%%%%%% ABSTRACT
\begin{abstract}
    Diffusion models have demonstrated impressive performance in text-guided image generation. Current methods that leverage the knowledge of these models for image editing either fine-tune them using the input image (\eg, Imagic) or incorporate structure information as additional constraints (\eg, ControlNet). However, fine-tuning large-scale diffusion models on a single image can lead to severe overfitting issues and lengthy inference time. Information leakage from pretrained models also make it challenging to preserve image content not related to the text input. Additionally, methods  that incorporate structural guidance (\eg, edge maps, semantic maps, keypoints) find retaining attributes like colors and textures difficult. Using the input image as a control could mitigate these issues,  but since these models are trained via reconstruction, a model can simply hide information about the original image when encoding it to perfectly reconstruct the image without learning the editing task. To address these challenges, we propose a text-to-image editing model with an Image Information Removal module (IIR) that selectively erases color-related and texture-related information from the original image, allowing us to better preserve the text-irrelevant content and avoid issues arising from information hiding. Our experiments on CUB, Outdoor Scenes, and COCO reports our approach achieves the best editability-fidelity trade-off results. In addition, a user study on COCO shows that our edited images are preferred $35$\% more often than prior work.
\end{abstract}

%%%%%%%%% BODY TEXT
\section{Introduction}
\label{sec:introduction}

\begin{figure}[!ht]
    \centering
    \includegraphics[width=1.\linewidth]{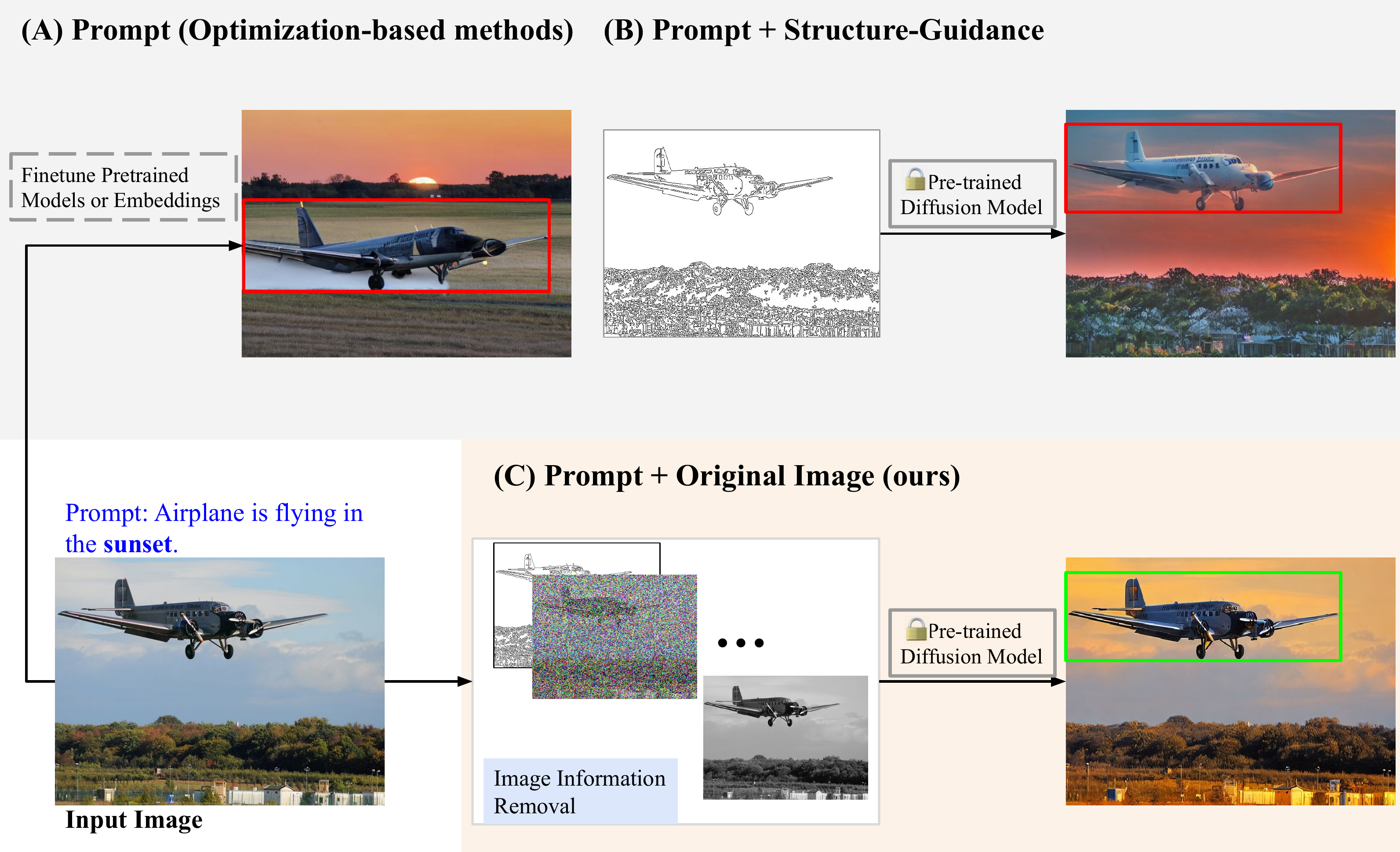}
    \caption{We aim to edit the specific content of the input image according to text descriptions while preserving text-irrelevant image content. Prior work based on large-scale diffusion models has followed two major approaches for image editing: (A), fine-tuning the pretrained models or text embeddings (\eg, Imagic~\cite{kawar2023imagic} or Dreambooth~\cite{ruiz2022dreambooth}), or (B), introducing structural guidance as additional constraint to control the spatial information of the generated image (\eg, ControlNet~\cite{zhang2023adding} or MaskSketch~\cite{bashkirova@masksketch}). In our work, shown in (C), our approach conditions on both the original image and the structural guidance, to better preserve the text-irrelevant content of the image. \Eg, our model successfully preserves the original attributes of the airplane (outlined by the green bounding box) in the generated image. In contast, previous methods such as Imagic (A) and ControlNet (B) not only alter the sky and background but also modify the attributes of the airplane (outlined by the red bounding boxes), which is unwanted in this example.
    }
\label{fig1:overview}
\end{figure}

Text-driven image editing aims to modify the specific content of an image based on its textual descriptions. Inspired by the powerful capability of large-scale text-to-image generation models~\cite{rombach2022high, ramesh2021zero,saharia2022photorealistic,li2023gligen}, recent approaches have leveraged the prior knowledge of these pretrained models for image editing~\cite{ruiz2022dreambooth, kawar2023imagic, bar2022text2live, zhang2022sine, zhang2023adding, bashkirova@masksketch}. The majority of existing editing approaches follow two strategies: 1)  Optimization-based methods: updating network parameters or feature embeddings for each input image, as shown in Figure \ref{fig1:overview} (A); or 2) introducing the structural guidance (\eg, edge map, user scribble, segmentation map, or pose estimation) as additional constraints for image generation, as shown in Figure \ref{fig1:overview} (B). The effectiveness of these models have been demonstrated on tasks like style transfer~\cite{zhang2022sine}, texture editing~\cite{bar2022text2live}, shape editing~\cite{kawar2023imagic}, appearance modification~\cite{ruiz2022dreambooth}, color editing~\cite{zhang2023adding}, among others. However, for optimization-based methods, fine-tuning large-scale models on single or few images results in severe over-fitting issues and prolongs inference time~\cite{zhang2022sine}. Images generated by finetuned models and embeddings may contain unexpected visual artifacts due to the information leakage and fail to preserve the text-irrelevant content of original image~\cite{zhang2022sine}. Structure-guided methods also meet pitfalls: structural guidance usually contains no information about colors or textures, these frameworks have difficulty preserving the text-irrelevant content of the original image. As outlined by red bounding boxes in Figure \ref{fig1:overview} (A) and (B), we observe both Imagic~\cite{kawar2023imagic} and ControlNet~\cite{zhang2023adding} fail to preserve the text-irrelevant content of the original image: Imagic modifies the shape of the airplane while ControlNet changes the color and textures of airplane. 

To address the aforementioned issues, we introduce the original image as an additional control for image editing. In this way the model can fully incorporate the content of the input image, allowing it to effectively preserve the text-irrelevant content. However, this results in an identity mapping issue~\cite{li2020manigan}, where the model can simply map the input directly to the output. This is primarily caused by the image reconstruction objective in editing task, which is perfectly optimized using an identity mapping. Prior works attempt to alleviate such issue by either learning disentangled features~\cite{locatello2019challenging, yang2021l2m, yao2021latent}, or uses attribute classifier to remove the target attribute~\cite{li2022supervised, li2022collecting}.  Both these approaches unavoidably introduce additional computational overhead that also greatly limits their application scenarios. For example, in Figure \ref{fig1:overview}, the input image only has text annotations and does not has scene attribute labels such as ``\textit{daylight}'' or ``\textit{sunset}.'' Therefore, these methods cannot be applied to convert the input image from ``\textit{daylight}'' to ``\textit{sunset}.'' 
% the editing models are typically trained on the image reconstruction task,

% To introduce the original image as input and avoid the identical mapping issue, 

We propose an Image Information Removal module (IIR-Net) to partially remove the image information from the input image, as illustrated in Figure \ref{fig1:overview} (C). Specifically, this erasure of image information arises from two components. First, we localize the Region of Interest (RoI\footnote{We refer to the modified regions of the target image as RoI. In our work, RoI is localized by Grounded-SAM~\cite{liu2023grounding,kirillov2023segment}. For tasks that the entire image is subject to modification such as scene attribute transfer or style transfer, we simply define the entire image as the RoI.}) and erase the color-related information. Second, we apply Gaussian noise on the input image which randomly eliminates the texture-related information. By tweaking the noise intensity, the model is capable of adapting to various tasks accordingly. For example, in color editing tasks, we decrease the noise intensity to zero to preserve most information from the input image except the color. In the texture editing task, a higher value of noise intensity is used to eliminate most information from the target region, leaving only the structural prior. Given the original image, we then concatenate the structure map with attribute-excluded features as additional controls to editing model. With our simple while effective image information removal module, we avoid the identical mapping issue as now the model is forced to not only reconstruct the original, but also predict the noised image regions. 
% the input to our model is different from the output, thus 
%In this case, our model is trained on rely on textual descriptions to reconstruct the missing information from the input image. 

We summarize the contribution of our work as follows:

\begin{itemize}[nosep,leftmargin=*]
    \item  We introduce the original image as an additional guidance to pretrained generative diffusion models for image editing tasks. Compared with existing image editing methods~\cite{kawar2023imagic, zhang2023adding}, IIR-Net more effectively preserves the text-irrelevant content of the input image while also generating new features according to the language descriptions. 
    \item We propose an image information removal module to counter the identical mapping issue~\cite{li2020manigan}. IIR-Net partially erases the image information such as colors or textures from the input, and reconstruct the original image according to text descriptions and attribute-excluded features. Compared with prior work for solving this issue~\cite{yang2021l2m,yao2021latent,li2022supervised}, IIR-Net does not require attribute labels to learn disentangled features or attribute classifiers, and, thus, can be applied to images without attribute labels. 
    \item We conduct extensive quantitative and qualitative experiments on three public datasets CUB~\cite{wah2011caltech}, COCO~\cite{lin2014microsoft}, and Outdoor Scenes~\cite{laffont2014transient}. Our results demonstrate that our model improves the fedility-editability trade-off over the state-of-the-art with obvious inference speed advantages. \Eg, compared to Imagic~\cite{kawar2023imagic}, IIR-Net improves the LPIPS score from $0.57$ to $0.30$ on COCO, with an inference speed improvement of two orders of magnitude.
\end{itemize}

\section{Related Work}
\label{sec:related_work}
\noindent \textbf{Feed-forward transformation image generation and editing.} 
Early work in text-to-image generation and editing often used text-to-image generator based on conditional GANs~\cite{reed2016generative, xu2018attngan, zhang2017stackgan, li2020manigan, li2020lightweight, nam2018text, dhamo2020semantic,tao2023net}. Limited by the scalability of Conditional GAN and size of image datasets, these methods only supported specific image domains and language descriptions. More recent methods typically trained conditional diffusion models~\cite{ramesh2021zero,ramesh2022hierarchical,rombach2022high,saharia2022photorealistic} on massive datasets (\eg, LAION-400M~\cite{schuhmann2021laion}). Due to the difficulty to obtain image pairs before and after editing, current image editing frameworks~\cite{zhang2023adding, bashkirova@masksketch, kawar2023imagic, ruiz2022dreambooth, zhang2022sine, nichol2021glide} are mostly developed based on pretrained text-to-generation models~\cite{rombach2022high, saharia2022photorealistic}. However, among these methods, methods that leverage the feed-forward transformation mechanism mostly focus on structural guidance. \Eg, ControlNet~\cite{zhang2023adding} leverages structure maps like edge map, semantic map, or pose estimation to control the spatial structure of generated images, and MaskSketch~\cite{bashkirova@masksketch} uses sketch as additional control to generate images. Thus, these methods cannot preserve the other attributes of the image such as colors or textures well, and may result in significant deviation from the input image. To solve this issue, we incorporate the original image as input to our model and propose an image information removal to solve the identical mapping issue~\cite{li2020manigan}.
\smallskip

\noindent \textbf{Optimization-based Methods} 
Prior work has demonstrated that optimization-based methods, which update network parameters on each image input, work well for image generation~\cite{shaham2019singan, tumanyan2022splicing, zhang2022semantic}. Several methods use CLIP~\cite{radford2021learning} as a constraint to guide the embedding features of predicted images~\cite{frans2022clipdraw, jain2022zero, michel2022text2mesh, bar2022text2live}. Inspired by the success of pretrained text-to-image generation frameworks~\cite{saharia2022photorealistic, ramesh2021zero, ramesh2022hierarchical}, researchers have also proposed methods to finetune these models for image editing (\eg, Imagic~\cite{kawar2023imagic}, Dreambooth~\cite{ruiz2022dreambooth}, SINE~\cite{zhang2022sine}, Textual Inversion~\cite{gal2022image}). Compared to feed-forward transformation methods~\cite{zhang2023adding, bashkirova@masksketch}, these models retain more information from the original image since they take the whole image instead of just a structure map as additional guidance. However, as we will show in Section \ref{sec:experiments_qualitative}, some image content such as background or irrelevant attributes of target objects may still be changed in this process. In addition, the inference time of these optimization-based methods is much longer than feed-forward transformation methods due to image-specific finetuning. %In this paper, we explore to combine the advantages of both optimization-based methods and feed-forward transformation methods. We incorporate the whole image as additional guidance to our model, addressing the information loss from the input image, and apply a feed-forward transformation mechanism to train our method, addressing the long inference time.

%To address these challenges, we build our model based on the feed-forward transformation and compositional scene representations. In this case, the content of original images can be preserved well and the inference process is faster than optimization-based methods.

\begin{figure*}[t]
    \centering
    \includegraphics[width=1.\linewidth]{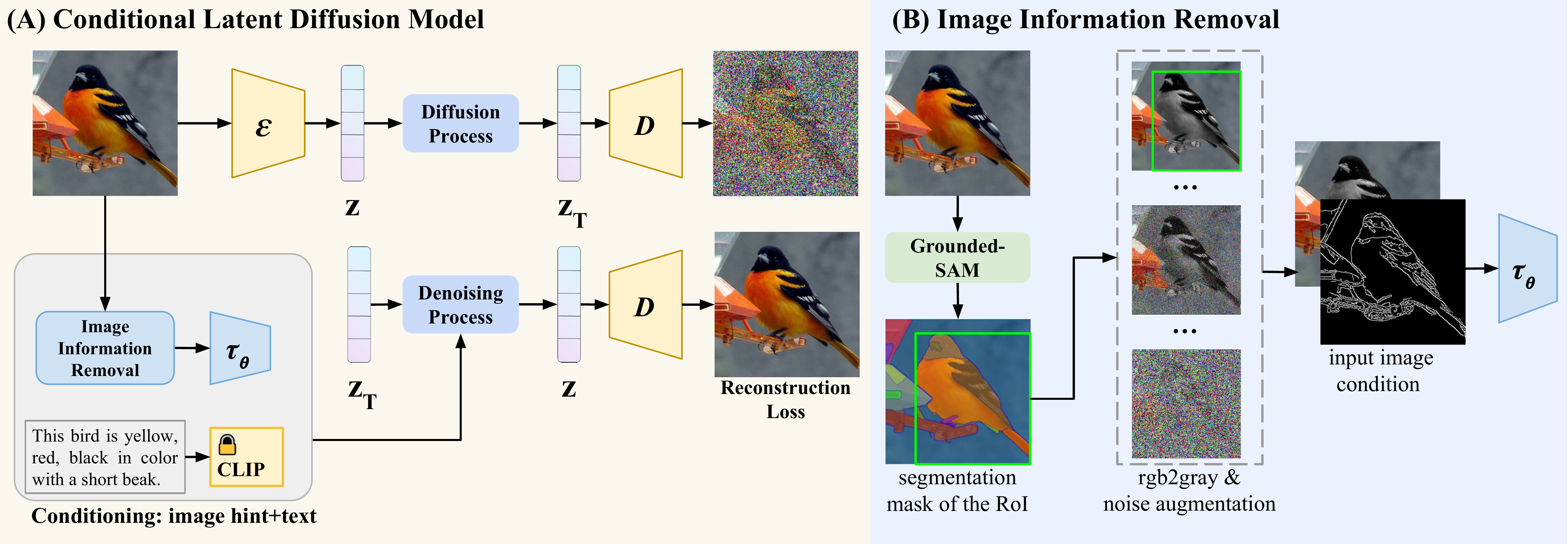}
    \caption{\textbf{IIR-Net Overview.} Our model mainly consists of two modules: (A) Conditional Latent Diffusion Model: We introduce the original image $x$ as additional control to our model to preserve the text-irrelevant features of $x$. See Section \ref{sec:IIR-Net_cldm} for detailed discussion; (B) Image Information Removal Module: We erase the image information mainly by two operations. First, we convert RGB values to the gray values in the RoI to exclude the color information. Second, we add Gaussian noise to the input image to partially erase the texture-related information. This module is applied to address the identical mapping issue. See Section \ref{sec:IIR-Net_removal_reconstruction} for detailed discussion.
    }
    \label{fig2:approach}
\end{figure*}

\section{IIR-Net: Text-to-Image Editing by Image Information Removal}
\label{sec:IIR-Net}

Given an input image $x$ and its corresponding text prompt $S$, our task aims to create the desired content according to $S$ while preserving the text-irrelevant content of $x$. To achieve this, we incorporate the original image $x$ as an additional control to pretrained text-to-image generation model, which is discussed in Section \ref{sec:IIR-Net_cldm}. However, since the model is trained on the image reconstruction task, the incorporation of the original image can lead to the identical mapping issue, in which the model simply maps the input image as the output. To address this challenge, we propose our image information removal module in Section \ref{sec:IIR-Net_removal_reconstruction}. Figure \ref{fig2:approach} provides an overview of our approach.

\subsection{Conditional Latent Diffusion Model}
\label{sec:IIR-Net_cldm}
As discussed in the Introduction, preserving the text-irrelevant content of the original image is critical for text-to-image editing. Leveraging the structural guidance as an additional hint (\eg, ControlNet\cite{zhang2023adding}, MaskSketch~\cite{bashkirova@masksketch}) can lead to significant information loss from the original image. To address this, we introduce the original image as additional control to our model, which preserves all information from the input image. In this section, we first introduce the pretrained text-to-image generation model, Stable Diffusion~\cite{rombach2022high}, as preliminaries to our method, and discuss our IIR component in Section~\ref{sec:IIR-Net_removal_reconstruction}.

Given an input image $x_0$ and its corresponding noisy image $x_T$, Stable Diffusion~\cite{rombach2022high} consists of a series of equally weighted denoising autoencoders $\epsilon_{\theta}(x_t,t)$, where $t$ ranges from $1\sim T$. The deonising autoencoders are trained to predict the noise $\epsilon$ in $x_T$ according to time step $t$ and noisy input $x_t$. The objective function of Stable Diffusion is 
\begin{align}
    L_{\rm{LDM}} := \mathbb{E}_{\mathcal{E}(x),\epsilon\sim \mathcal{N}(0,1),t}\left[ || \epsilon - \epsilon_\theta(z_t,t) ||^2_2 \right],
\label{eq:ldm}
\end{align}
\noindent where $\mathcal{E}$ is the pretrained encoder of VQGAN~\cite{esser2021taming} to encode image $x_t$ to latent features $z_t$, and vice versa. For conditional generation, the denoising autoencoders $\epsilon$ take $\tau_{\theta}(y)$ as additional input, where $\tau_{\theta}(y)$ represents a domain-specific encoder to extract feature embeddings from the condition $y$.  This condition $y$ represents elements like text prompts and semantic maps, among others. Given image-condition pairs, the Conditional Latent Diffusion Model (CLDM) is optimized by
\begin{align}
    L_{\rm{CLDM}} := \mathbb{E}_{\mathcal{E}(x), y, \epsilon\sim \mathcal{N}(0,1),t}\left[ || \epsilon - \epsilon_\theta(z_t,t, \tau_{\theta}(y)) ||^2_2 \right],
\label{eq:conditional_ldm}
\end{align}
\noindent where $\tau_\theta$, $\epsilon_\theta$ are jointly optimized. In our model, the condition $y$ consist of text descriptions $S$ and the original image $x_0$ and is defined as
\begin{align}
    \tau_{\theta}(y) := \{\tau_{\theta_1}(S), \tau_{\theta_2}(R(x_0))\},
\label{eq:condition}
\end{align}
where we use the CLIP model~\cite{radford2019language} as $\tau_{\theta_1}(\cdot)$ to encode the text descriptions $S$ and use ControlNet~\cite{zhang2023adding} as $\tau_{\theta_2}(\cdot)$ to encode the feature $R(x_0)$. $R(\cdot)$ denotes our image information removal module, which we discuss in the next section.

\subsection{Image Information Removal}
\label{sec:IIR-Net_removal_reconstruction}
As discussed in the Introduction, training solely on image reconstruction can lead to the identical mapping issue. Previous approaches address this issue by learning disentangled features~\cite{gal2022stylegan} or attribute classifiers~\cite{li2022supervised}. However, these methods require annotated attributes, restricting their application scenarios. To overcome this challenge, we propose our image information removal module, which incorporates color and texture removal operations. Our removal operations effectively mitigates the identical mapping issue without the requirement for additional annotated labels.
\smallskip

\noindent \textbf{Color-related Information Removal.} In Figure \ref{fig2:approach} (B), we present our color information removal operation. Given the input image $x_0$ and its corresponding text prompt $S$, we employ Grounded-SAM~\cite{liu2023grounding,kirillov2023segment} to localize the RoI. The color information of $x_0$ is then erased by 
\begin{align}
    x_0' = \rm{rgb2gray}(x\odot m_{RoI})+x\odot(1-m_{RoI}),
\label{eq:erase_color}
\end{align}
where $m_{RoI}$ is the Grounded-SAM segmentation mask. 
%\smallskip

Through the application of color-related information removal to the input image $x_0$, our model demonstrates proficiency in color-related editing tasks, such as transforming a "white airplane" into a "green airplane." However, as depicted in Figure \ref{fig:ablation}, the model encounters challenges when attempting to modify texture-related information, such as changing "lawn" to "sand." To address this limitation, we introduce our texture-related information removal module.
\smallskip

\noindent \textbf{Texture-related Information Removal.} We eliminate the texture-related information by adding noise to the image condition $x_0'$ of our model, denoted by
\begin{align}           
    q(x_K'|x_0')&=\prod_{k=1}^{K}q(x_k'|x_{k-1}');\\
    \hspace{4mm} q(x_k'|x_{k-1}') &= \mathcal{N}(x_k'; \sqrt{1-\beta_k}x_{k-1}'; \beta_k \mathbf{I}),
\label{eq:erase_texture}
\end{align}
where $k$ denotes the time step applied to $x_{k}'$, which is different from the time step $t$ applied to $x_t$. Note that $x_t$ is obtained by adding noise to the original image $x_0$ in diffusion models, whereas $x_k'$ is obtained by adding noise to the image condition $x_0'$ in diffusion models. During training we randomly sample $x_k'$ from $\{x_0',\ldots,x_K'\}$.

While $x_k'$ inherently preserves the structure information of $x_0$, we find that explicitly incorporating additional structural guidance, such as edges, helps the model better capture structural information. Thus, we concatenate $x_k'$ with the predictions of a Canny Edge detector $\mathbf{C}(x_0)$. Thus, the output of our image information removal module is:  
\begin{align}
    R(x_0) = \left[{x_k',\mathbf{C}(x_0)}\right].
\label{eq:removal_output}
\end{align}

Given the output of our image information removal module $R(x_0)$, the final objective of IIR-Net is defined as:
% \begin{equation}
%     L_{\rm{IIR-Net}} := \mathbb{E}_{\mathcal{E}(x), y, \epsilon\sim \mathcal{N}(0,1),t}\left[ || \epsilon - \epsilon_\theta(z_t,t, \tau_{\theta_1}(S),\\ \tau_{\theta_2}(R(x_0))) ||^2_2 \right].
% \label{eq:iir-net}
% \end{equation}
\begin{equation}
\begin{aligned}
    L_{\rm{IIR-Net}} := \mathbb{E}_{\mathcal{E}(x), y, \epsilon\sim \mathcal{N}(0,1),t}\Big[ \| \epsilon - \epsilon_\theta(z_t,t, \tau_{\theta_1}(S), \\
    \tau_{\theta_2}(R(x_0))) \|^2_2 \Big].
\end{aligned}
\label{eq:iir-net}
\end{equation}

\section{Experiments}
\label{sec:experiments}

\subsection{Datasets and Experiment Settings}
\noindent \textbf{Datasets.} We evaluate the performance of our model on three standard datasets, CUB~\cite{wah2011caltech}, Outdoor Scenes~\cite{laffont2014transient}, and COCO~\cite{lin2014microsoft}. CUB~\cite{wah2011caltech} is contains 200 bird species that we split into 8,855 training images and 2,933 test images. Ourdoor Scenes~\cite{laffont2014transient} contains 8,571 images captured from 101 webcams, with each webcam collecting 60$\sim$120 images showcasing different attributes like weather, season, or time of day. COCO~\cite{lin2014microsoft} contains 82,783 training images and 40,504 validation images. Following~\cite{kawar2023imagic}, we randomly select 150 test images from each dataset to evaluate the performance of each method.
\smallskip

\noindent \textbf{Metrics.} Following~\cite{kawar2023imagic}, we adopt the perceptual metric LPIPS~\cite{zhang2018unreasonable} and CLIP score~\cite{radford2021learning} as our quantitative metrics. LPIPS measures the image fidelity and CLIP evaluates the model's editability. Additionally, we perform quantitative experiments by user study and inference time to evaluate the effectiveness and efficiency of our model.
\smallskip

\noindent \textbf{Baselines.} We compare IIR-Net with three state-of-the-art approaches: Text2LIVE~\cite{bar2022text2live}, Imagic~\cite{kawar2023imagic}, and ControlNet~\cite{zhang2023adding}. For Text2LIVE, we set the optimization steps to 600. For Imagic, both the text embedding optimization steps and model fine-tuning steps are set to 500. We sample the interpolation hyperparameter $\eta$ from 0.1 to 1 with a 0.1 interval, and the guidance scale is set to 3. For ControlNet and IIR-Net, we generate images with a CFG-scale of 9.0, and DDIM steps of 20 by default. %In our paper, our primary focus is on comparing IIR-Net with methods that leverage large-scale pretrained diffusion models. To produce high-quality and detailed images, we have omitted the comparisons to earlier conditional GAN-based approaches, such as~\cite{li2020manigan,li2020lightweight,tao2023net}. 
\smallskip

\begin{figure*}[t]
    \centering
    \includegraphics[width=1.0\linewidth]{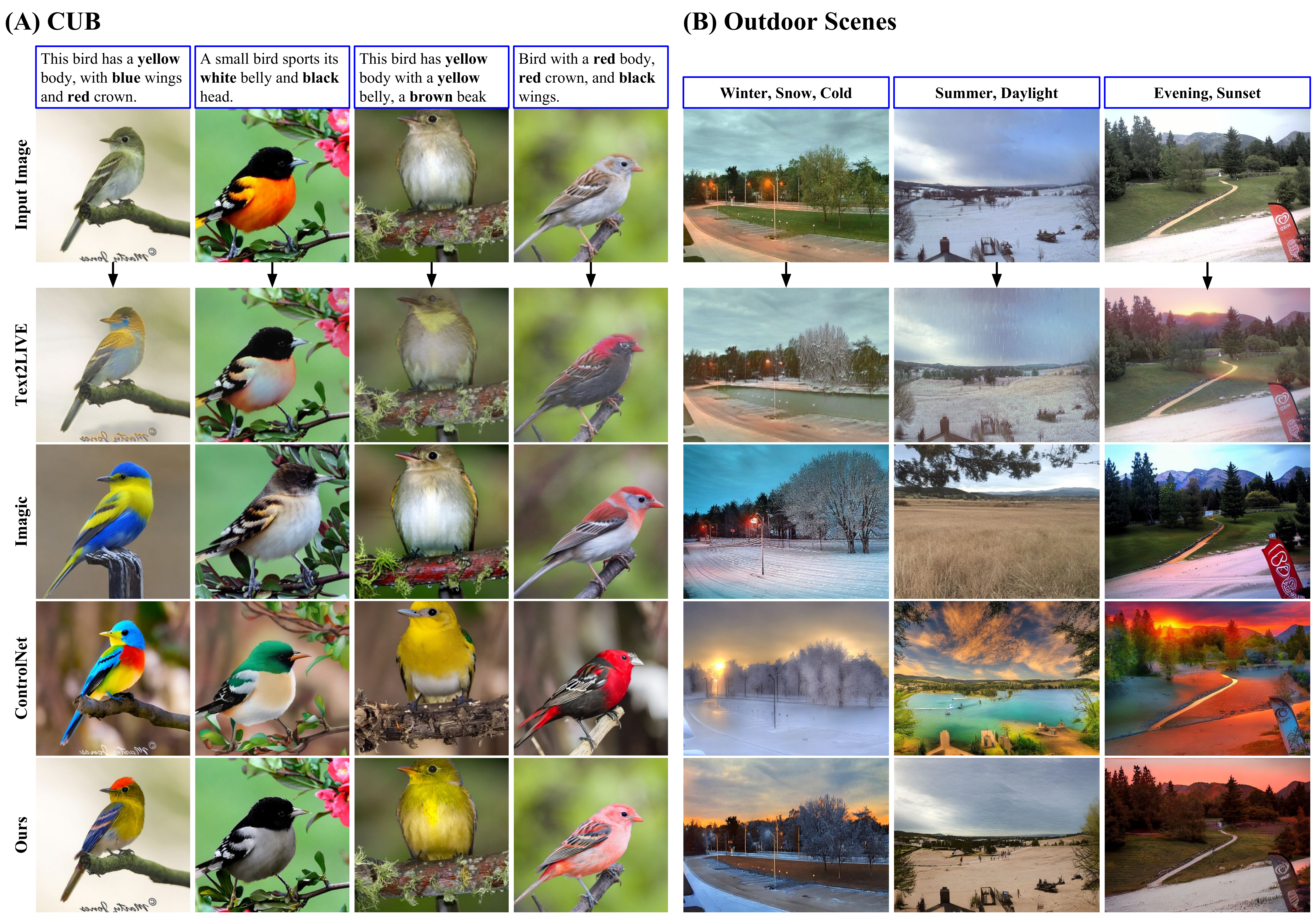}
    \vspace{-2mm}
    \caption{Qualitative comparison on CUB and Ourdoor Scenes. From top to bottom: input image, Text2LIVE~\cite{bar2022text2live}, Imagic~\cite{kawar2023imagic},  ControlNet~\cite{zhang2023adding}, and ours. Generated images have 512 pixels on their shorter side. See Section \ref{sec:experiments_qualitative} for discussion.}
    \label{fig:cub_scene_qualitative}
    \vspace{-2mm}
\end{figure*}

\noindent \textbf{Implementation Details.}
We initialized our model weights from Stable Diffusion 1.5~\cite{rombach2022high} and ControlNet~\cite{zhang2023adding}. During training, we applied a batch size of 8 and a maximum learning rate of $1\times10^{-6}$. We finetuned our models approximately 100 epochs on the CUB~\cite{wah2011caltech} dataset, and around 5 epochs on the Outdoor Scenes~\cite{laffont2014transient} and COCO~\cite{lin2014microsoft} datasets. The training process was parallelized on 2 NVIDIA RTX-A6000s. To adapt the image conditions in our model, we configured the channel of the image encoder block to 4, with 3 channels for RGB images and 1 channel for the edge map. We finetuned the Stable Diffusion decoder for experiments on CUB, as these images primarily focus on various birds with a consistent style. We froze the Stable Diffusion Decoder for the Ourdoor Scenes and COCO datasets, since these datasets comprising natural images with diverse objects and varying styles. %Following~\cite{zhang2023adding}, during inference we set the CFG-scale to 9 and the diffusion steps to 20 by default.

% \noindent \textbf{Implementation Details}
% The image information removal module (Section \ref{sec:IIR-Net_removal_reconstruction}) consists of two operations, masking image and RGB-to-gray conversion. Different from representation learnings~\cite{he2022masked} that mask a high proportion of the input image, we cropped out image patches to avoid identity mapping while still would like to preserve enough information for image editing. Therefore, the masking ratio of our model is set to 0.3. For latent diffusion module (Section \ref{sec:IIR-Net_cldm}), we build our method using the same architecture of Stable Diffusion~\cite{rombach2022high}. To keep the dimension of image embeddings consistent with the dimension of text embeddings (768), we extract image features using the first four layers of CLIP's ViT encoder. During training, the parameters of CLIP~\cite{radford2021learning} encoders and VQGAN~\cite{esser2021taming} encoders are fixed, we only update the parameters of the latent diffusion model. We apply a batch size of 8 and a maximum learning rate of $1\times10^{-6}$. We trained our model around 200 epochs on CUB and 100 epochs on COCO with 1,000 linear warm-up on both datasets. During inference, we set the unconditional guidance scale to 9 and the diffusion step to 20. 

\begin{figure*}[t]
    \centering
    \includegraphics[width=0.9\linewidth]{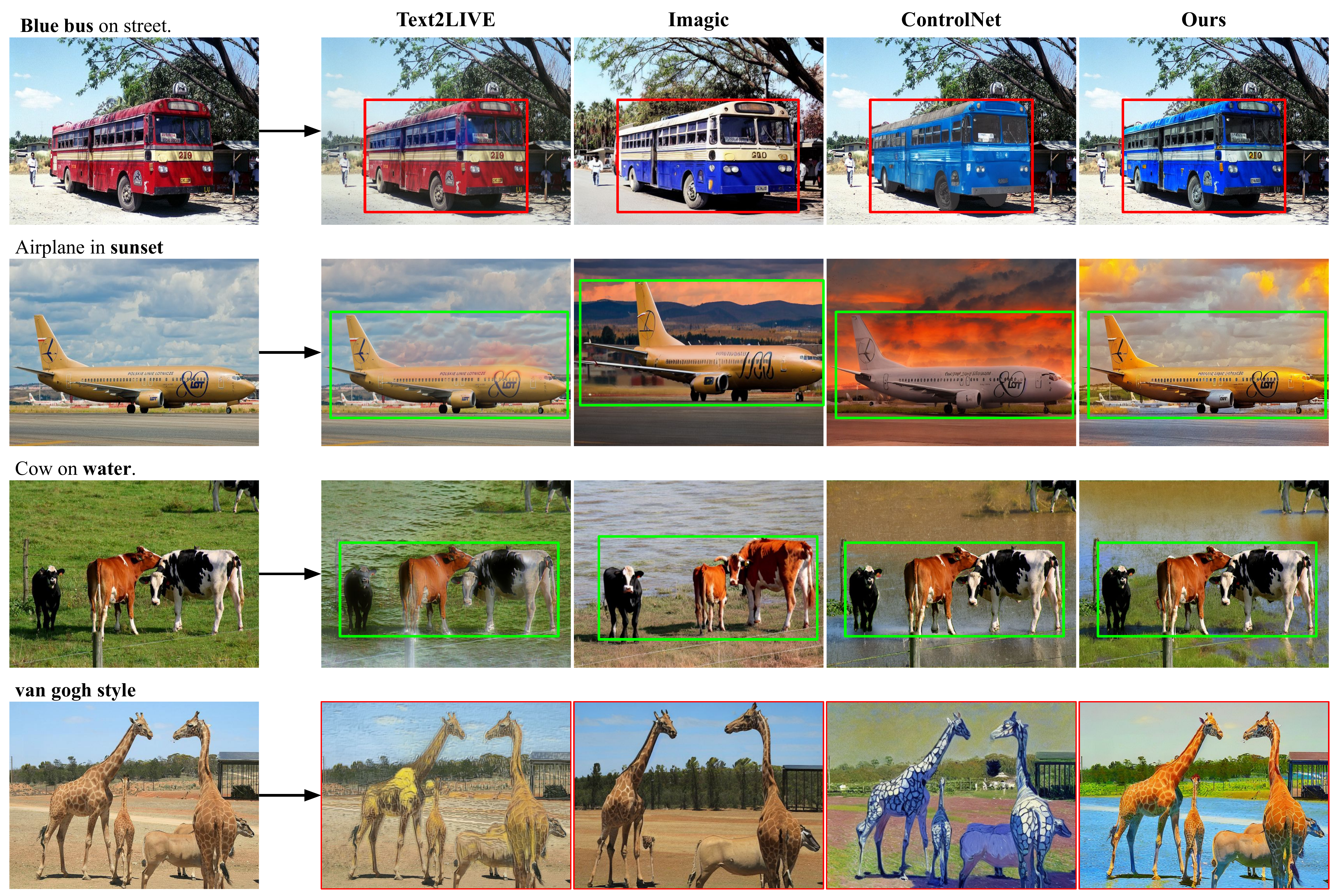}
    \vspace{-2mm}
    \caption{Qualitative comparison for various editing tasks on the COCO dataset. From top to bottom: color editing, scene attribute transfer, texture editing, and style transfer. Generated images have 512 pixels on their shorter side. Objects that are the target of modification are in red bounding boxes and whereas objects that should be preserved are in green bounding boxes. See Section \ref{sec:experiments_qualitative} for discussion.
    }
    \vspace{-4mm}
    \label{fig:coco_qualitative}
\end{figure*}

% \begin{table}
% \centering
% \setlength{\tabcolsep}{3.pt}
% \begin{tabular}{rlcc rlcc rlcc}
% \toprule
% \textbf{(a)} & \textbf{CUB} & LPIPS $\downarrow$ & CLIP $\uparrow$ & \textbf{(b)} & \textbf{Outdoor Scenes} & LPIPS $\downarrow$ & CLIP $\uparrow$ & \textbf{(c)} & \textbf{COCO} & LPIPS $\downarrow$ & CLIP $\uparrow$\\
% \midrule
% % & baseline1 & & & & baseline1 & & \\
% % & baseline2 & & & & baseline2 & & \\
% & Imagic~\cite{kawar2023imagic} & 0.406 & 27.03 & 
% & Imagic~\cite{kawar2023imagic} & 0.551 & 22.85 & 
% & Imagic~\cite{kawar2023imagic} & 0.567 & 21.53 \\

% & Text2live~\cite{bar2022text2live} & 0.162 & \textbf{30.37} & 
% & Text2live~\cite{bar2022text2live} & \textbf{0.218} & 22.64 & 
% & Text2live~\cite{bar2022text2live} & 0.495 & \textbf{25.11} \\

% & ControlNet~\cite{zhang2023adding} & 0.528 & 29.49 & 
% & ControlNet~\cite{zhang2023adding} & 0.618 & 23.89 & 
% & ControlNet~\cite{zhang2023adding} & 0.606 & 23.57 \\

% & ours & \textbf{0.138} & 29.57 & 
% & ours & 0.479 & \textbf{25.45} &
% & ours & \textbf{0.301} & 24.30\\

% \bottomrule
% \end{tabular}
% % \vspace{0mm}
% \caption{Quantitative experiments of image manipulation on CUB~\cite{wah2011caltech}, Outdoor Scene~\cite{laffont2014transient}, and COCO~\cite{lin2014microsoft} datasets. CLIP is used to evaluate the image editing performance and LPIPS is applied to evaluate image fidelity. See Section \ref{sec: experiment_quantitative} for discussion.}
% \label{table:quantatitive}
% \end{table}

\subsection{Qualitative Results}
\label{sec:experiments_qualitative}

\noindent \textbf{Entire-image Editing on the CUB and Outdoor Scenes Datasets.} 
Figure \ref{fig:cub_scene_qualitative} presents a qualitative comparison of the edited images generated by our model and the baselines. In Figure \ref{fig:cub_scene_qualitative} (A), we present a comparison on the CUB~\cite{wah2011caltech} dataset. We observe that our model can accurately manipulate parts of the bird while preserving the text-irrelevant content of the original image. For example, in the first column of Figure \ref{fig:cub_scene_qualitative} (A), while baselines such as ControlNet and Imagic can recognize ``yellow'' and ``blue'' from the text prompt, they both fail to effectively apply them to the correct corresponding parts of the bird. Imagic generates a bird with a blue crown and yellow wings, while ControlNet generates a blue head and a red breast. In constrast, our model accurately edits the bird by parts according to the prompt and produce a bird with blue wings, yellow body, and red crown. In addition, we observe that the background of images generated by Imagic and ControlNet has been changed. This is due to the fact that Imagic and ControlNet do not directly use the original image as their input. \Eg, Imagic optimizes the text embeddings to get features that reflect the attributes of the original image, and ControlNet uses the Canny Edge map as input. Thus, it is challenging for these method to preserve the text-irrelevant content of the original image. In contrast, our model takes the original image as input and only erases the text-relevant content, thus preserving the text-irrelevant content effectively.

In Figure \ref{fig:cub_scene_qualitative} (B), we present a comparison on the Outdoor Scenes~\cite{laffont2014transient} dataset. Consistent with our findings on the CUB dataset, we observe that baselines like Imagic and ControlNet tend to modify the text-irrelevant contents of the original image, such as the textures and background, while Text2LIVE only introduces limited visual effects to the original image and may fail to generate images aligned with the text descriptions. For example, in the second column of Figure \ref{fig:cub_scene_qualitative} (B), images produced by Imagic and ControlNet are well aligned with text descriptions (``summer,'' ``daylight''), but they introduce unexpected objects such as trees or a lake to the image. In contrast, Text2LIVE preserves the original image well, but fails to align with text descriptions, as seen with the snow-covered field in summer. However, our method effectively modifies the desired content, such as changing ``winter" to ``summer," while preserving the original content of the image.
\smallskip

\noindent \textbf{Region-based Image Editing on COCO.}
Unlike object-centric datasets such as CUB and Outdoor Scenes, COCO images can contain complex scenes with many objects, yet only parts of the input image may require modification. Thus, we apply Grounding-DINO~\cite{liu2023grounding} and SAM~\cite{kirillov2023segment} to localize the Region of Interest (RoI) that requires editing\footnote{Since Text2LIVE and Imagic automatically localize the RoI, we apply Grounding-DINO and SAM to ControlNet and our method.}. 

\begin{figure*}[t]
    \centering
    \includegraphics[width=0.99\linewidth]{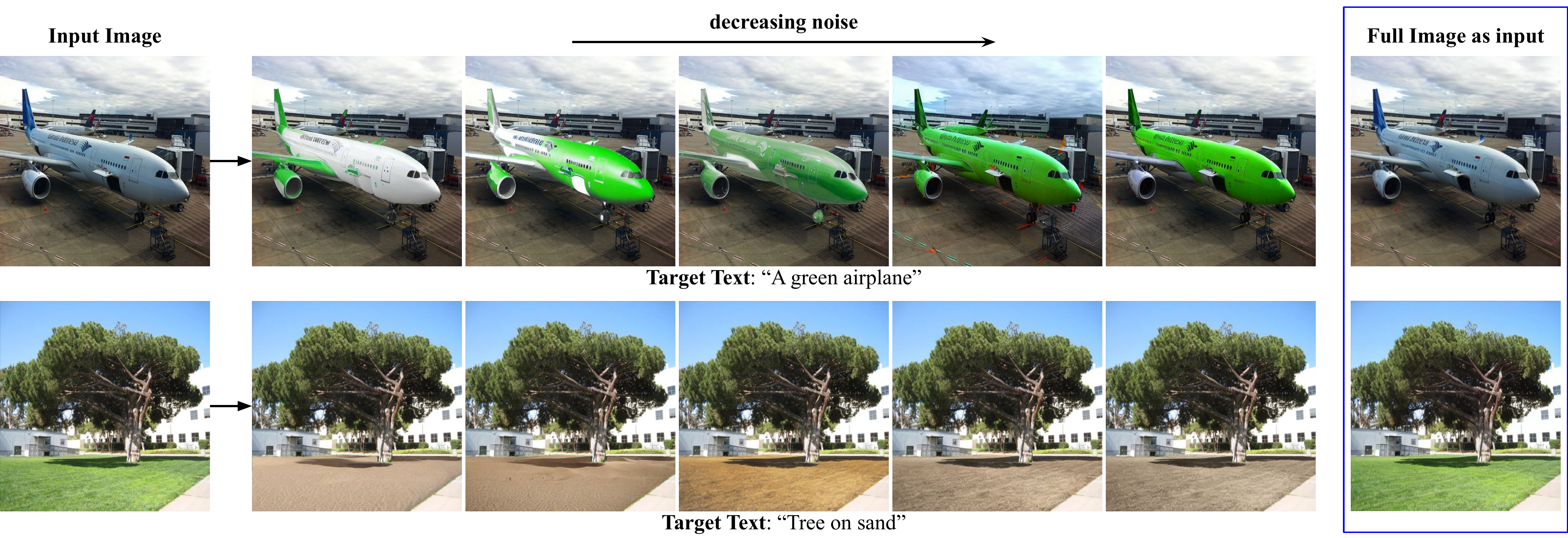}
    \vspace{-2mm}
    \caption{\textbf{Ablation Study.} We perform experiments to evaluate the effectiveness of our color removal and texture removal operations. Images generated without our image information removal module are outlined by the blue bounding box. See Section \ref{sec:experiments_qualitative} for discussion.}
    \vspace{-4mm}
    \label{fig:ablation}
\end{figure*}

 Figure \ref{fig:coco_qualitative} presents a qualitative comparison of our method and prior work on various image editing tasks. We find that our method produces images that are well-aligned with the text descriptions while non-edited components better represent the original images. \Eg, in the color editing task, although Imagic and ControlNet generate a blue bus according to the text prompt, Imagic changes the original shape of the bus and ControlNet modifies the bus' texture. In contrast, our method only modifies the color attribute while preserving irrelevant attributes. Furthermore, our model generates images that appear more natural and visually appealing. \Eg, in the scene attribute transfer task, the visual effect of ``sunset'' brought by our model is naturally aligned with the original image, whereas Text2LIVE introduces obvious artificial effects to the airplane. 

Finally, we evaluate our model on tasks where the original ControlNet performs well, such as texture editing and style transfer. Our results show that adapting text-to-image generation models to image editing tasks does not notably compromise their capabilities. For example, in the style transfer examples on COCO images our method still retains the ability to transfer a photorealistic image to an artistic style. See the supplementary for additional examples.
\smallskip

\noindent \textbf{Ablation Study.} 
In Figure \ref{fig:ablation}, we provide ablation study of IIR-Net. We find that without our unsupervised image content removal mechanism, the model always outputs the input image as the predicted image, \ie, the identical mapping issue~\cite{li2020manigan}. \Eg, the images in the blue bounding box remain white airplane and green grass, showing a lack of alignment with the text descriptions. By incorporating the color removal mechanism (see images with low noise level), our model performs well on tasks such as color editing. For example, when changing the airplane's color from white to green, our model preserves the most of the airplane's attributes, only modifying the color. We observe that the color removal mechanism can find texture editing challenging. For example, as seen in the second row of the figure, the images generated with low noise level still exhibit the grass texture instead of the intended ``sand'' texture. Therefore, we incorporate noise augmentation to the input images to better handle such editing tasks. As shown in the second row, our model successfully modifies the grass texture to sand under high-level noise conditions. In practical applications, users can adjust the noise level according to different editing tasks to achieve optimal performance.

\subsection{Quantitative Results}
\label{sec:experiments_quantitative}

% \begin{table}[t]
% \centering
% \setlength{\tabcolsep}{2pt}
% \begin{tabular}{rlcc rlcc rlcc}
% \toprule
% \textbf{(A)} & \textbf{CUB} & LPIPS $\downarrow$ & CLIP $\uparrow$ & \textbf{(B)} & \textbf{Scenes} & LPIPS $\downarrow$ & CLIP $\uparrow$ & \textbf{(C)} & \textbf{COCO} & LPIPS $\downarrow$ & CLIP $\uparrow$\\
% \midrule
% % & baseline1 & & & & baseline1 & & \\
% % & baseline2 & & & & baseline2 & & \\
% & Imagic~\cite{kawar2023imagic} & 0.406 & 27.03 & 
% & Imagic~\cite{kawar2023imagic} & 0.551 & 22.85 & 
% & Imagic~\cite{kawar2023imagic} & 0.567 & 21.53 \\

% & Text2live~\cite{bar2022text2live} & 0.162 & \textbf{30.37} & 
% & Text2live~\cite{bar2022text2live} & \textbf{0.218} & 22.64 & 
% & Text2live~\cite{bar2022text2live} & 0.495 & \textbf{25.11} \\

% & ControlNet~\cite{zhang2023adding} & 0.528 & 29.49 & 
% & ControlNet~\cite{zhang2023adding} & 0.618 & 23.89 & 
% & ControlNet~\cite{zhang2023adding} & 0.606 & 23.57 \\

% & ours & \textbf{0.138} & 29.57 & 
% & ours & 0.479 & \textbf{25.45} &
% & ours & \textbf{0.301} & 24.30\\

% \bottomrule
% \end{tabular}
% \caption{Quantitative experiments of image manipulation on CUB~\cite{wah2011caltech}, Outdoor Scene~\cite{laffont2014transient}, and COCO~\cite{lin2014microsoft} datasets. CLIP~\cite{radford2021learning} is used to evaluate the image editing performance and LPIPS is applied to evaluate image fidelity. Generated images have been resized to 224$\times$224 resolution for CLIP score. We use the ``ViT-B/32'' version of CLIP. See Section \ref{sec:experiments_quantitative} for discussion.}
% \label{table:quantatitive}
% \end{table}

\begin{table*}[t]
\centering
\setlength{\tabcolsep}{4pt}
\begin{tabular}{l rcc rcc rcc}
\toprule
&   & \multicolumn{2}{c}{\textbf{CUB}} &   & \multicolumn{2}{c}{\textbf{Outdoor Scenes}} &   & \multicolumn{2}{c}{\textbf{COCO}} \\ 
% \cline{2-10}

& & LPIPS $\downarrow$ & CLIP $\uparrow$ & & LPIPS $\downarrow$ & CLIP $\uparrow$ & & LPIPS $\downarrow$ & CLIP $\uparrow$\\
\midrule
% & baseline1 & & & & baseline1 & & \\
% & baseline2 & & & & baseline2 & & \\
Imagic~\cite{kawar2023imagic} & & 0.406 & 27.03 & 
 & 0.551 & 22.85 & 
 & 0.567 & 21.53 \\

Text2live~\cite{bar2022text2live} & & 0.162 & \textbf{30.37} & 
 & \textbf{0.218} & 22.64 & 
 & 0.495 & \textbf{25.11} \\

ControlNet~\cite{zhang2023adding} & & 0.528 & 29.49 & 
 & 0.618 & 23.89 & 
& 0.606 & 23.57 \\

ours & & \textbf{0.138} & 29.57 & 
 & 0.479 & \textbf{25.45} &
 & \textbf{0.301} & 24.30\\

\bottomrule
\end{tabular}
% \vspace{-2mm}
\caption{Quantitative experiments of image manipulation on CUB~\cite{wah2011caltech}, Outdoor Scenes~\cite{laffont2014transient}, and COCO~\cite{lin2014microsoft} datasets. CLIP~\cite{radford2021learning} is used to evaluate the image editing performance and LPIPS is applied to evaluate image fidelity. Generated images have been resized to 224$\times$224 resolution for CLIP score. We use the ``ViT-B/32'' version of CLIP. See Section \ref{sec:experiments_quantitative} for discussion.}
\vspace{-2mm}
\label{table:quantatitive}
\end{table*}

\noindent \textbf{Editability-fidelity Tradeoff.}
Table \ref{table:quantatitive} reports our quantitative results on CUB, Outdoor Scenes, and COCO. As observed in our qualitative experiments, our model achieves a better tradeoff between image fidelity and editability compared to other state-of-the-art methods. \Eg, our model achieves the best LPIPS scores (0.138 and 0.301) and comparable CLIP scores (29.57 and 24.30) on CUB and COCO. In Outdoor Scenes, our model achieves the highest CLIP score and the second best LPIPS score. Text2LIVE achieves better LPIPS score than our method on Outdoor Scenes. However, it may due to the fact that Text2LIVE mainly augment the scenes with new visual effects, rather than directly modifying the attributes of the scenes. \Eg, Text2LIVE fails to change the grassland to a snowy landscape or convert lush trees to bare ones in the scenes. %Besides, as demonstrated in Figure \ref{fig:cub_scene_qualitative} (A), we notice that CLIP may not be fine-grained enough to evaluate the subtle details such as ``white belly'' or ``red crown.'' Therefore, the editability of our model may be under-reported. 
\smallskip

\noindent \textbf{User Study.} We conducted a user study to quantitatively evaluate the performance of IIR-Net, as shown in Table \ref{table:inference_time}. We randomly selected 30 images from COCO and applied each model to generate the modified images, resulting in a total of 120 generated images. Each image was annotated three times by users and we asked our annotators to judge whether the image is correctly manipulated based on the text guidance while preserving the text-irrelevant content of the original image. In the table, we report that IIR-Net significantly outperforms baselines. See the supplementary for additional details on our user study. 
\smallskip

\noindent \textbf{Inference Time}
Table \ref{table:inference_time} presents a comparison of the inference time and their standard error using the same Stable Diffusion v1.5~\cite{rombach2022high} backbone for Imagic, ControlNet, and our method. All methods are benchmarked on a NVIDIA RTX A6000 GPU. We find our method has significantly faster inference times compared to Imagic, boosting inference speed by two orders of magnitude when processing 512$\times$512 images. In addition, our method is approximately 50x faster than Text2LIVE. We note both ControlNet and our method have around 5s inference time, demonstrating that approach introduces negligible overhead to ControlNet.

\begin{figure}[t!]
    \centering
    \includegraphics[width=1.0\linewidth]{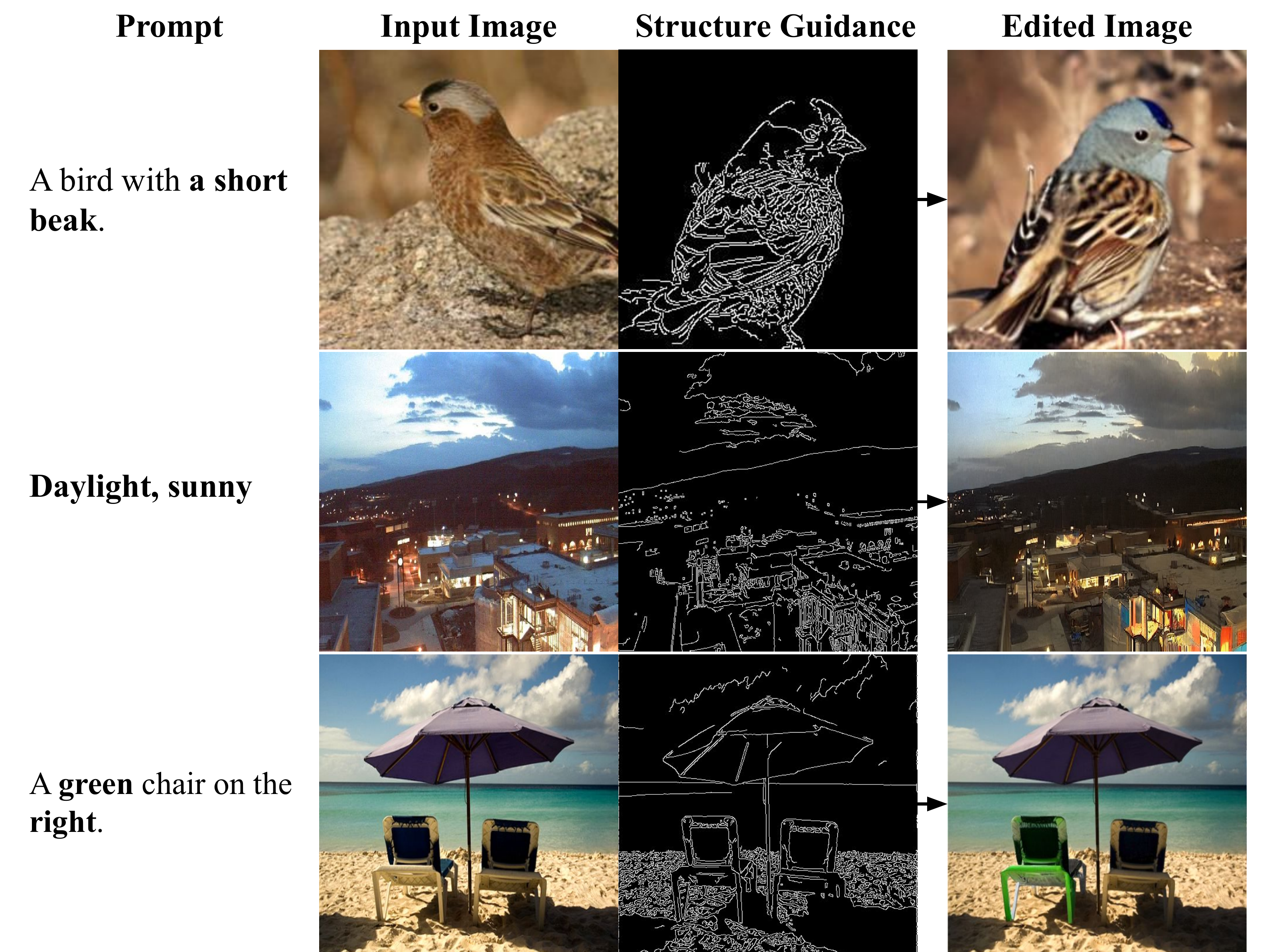}
    \vspace{-2mm}
    \caption{\textbf{Failure cases} include inconsistencies with the original image in non-rigid image editing task (top); challenges in notably modifying the brightness of the image (middle), and inaccurate localization or segmentation (bottom). See Section \ref{sec:experiments_limitation} for discussion.}
    % \vspace{-2mm}
    \label{fig:failure_case}
\end{figure}

% \begin{table}[!t]
% \centering
% % \setlength{\tabcolsep}{3.pt}
% \begin{tabular}{lcccc}
% \toprule
% Method & Text2LIVE~\cite{bar2022text2live} & Imagic~\cite{kawar2023imagic} & ControlNet~\cite{zhang2023adding} & IIR-Net (ours) \\
% \midrule
% User Preference & 30.0\% & 23.3\% & 33.3\% & \textbf{68.3\%} \\ 
% Inference Time & 281.6$\pm$1.72s & 483.4$\pm$1.31s & 5.0$\pm$0.04s & 5.0$\pm$0.03s  \\
% \bottomrule
% \end{tabular}
% \caption{We randomly select 30 images from COCO for user study and speed evaluation. Top row reports user judgments on the correctness of the image manipulation. Bottom row reports speed for our method v.s. the baselines. Our method has negligible overhead compared to ControlNet, and is significantly faster than Text2LIVE and Imagic. See Section \ref{sec:experiments_quantitative} for discussion.}
% \label{table:inference_time}
% \end{table}

\begin{table}[!t]
\centering
\begin{tabular}{lcc}
\toprule
Method & User Preference & Inference Time \\
\midrule
Text2LIVE~\cite{bar2022text2live} & 30.0\% & 281.6$\pm$1.72s \\
Imagic~\cite{kawar2023imagic} & 23.3\% & 483.4$\pm$1.31s \\
ControlNet~\cite{zhang2023adding} & 33.3\% & 5.0$\pm$0.04s \\
IIR-Net (ours) & \textbf{68.3\%} & 5.0$\pm$0.03s \\
\bottomrule
\end{tabular}
\caption{We randomly select 30 images from COCO for user study and speed evaluation. Top row reports user judgments on the correctness of the image manipulation. Bottom row reports speed for our method v.s. the baselines. Our method has negligible overhead compared to ControlNet, and is significantly faster than Text2LIVE and Imagic. See Section \ref{sec:experiments_quantitative} for discussion.}
\vspace{-2mm}
\label{table:inference_time}
\end{table}

\section{Limitations \& Broader Impacts.}
\label{sec:experiments_limitation}
\noindent \textbf{Limitations.} We identify three failure cases of our methods in this section: First, the attributes of the original image are likely to be modified in non-rigid image editing tasks. Second, it is challenging for our method to change the brightness of the input image drastically. Third, the target object may be localized and segmented inaccurately. We present examples of these three failure cases in Figure \ref{fig:failure_case}. As shown in the top row, though our method can achieve non-rigid image editing according to the input image and a modified structural guidance, we observe that the model fails to map some attributes to the correct parts. \Eg, the bird of the input image has a grey crown while the edited image generate a bird whose head is gray. The color of wings is also slightly different from the input bird. In the middle row, we find that our model fails to change the brightness of the image in some cases. \Eg, the input image is a night view. Therefore, the brightness of the image is low in this image and the model tend to reconstruct an image with a low brightness even if the target text is ``daylight,'' ``sunny.'' In the bottom row, we observe that our segmentation module fails to accurately localize the target object according to the prompt due to text ambiguity. While the prompt specifies the chair on the right-hand side, our model modifies the attributes of the chair on the left-hand side.
\smallskip

\noindent \textbf{Broader Impacts.}
Our model is designed to perform image editing according to user-provided language descriptions. Thus, it enables modification of attributes such as colors, textures, or styles in the original images. As other image generation and editing approaches, our model may be used to synthesize images that contains misinformation. Therefore, it is important for practitioners to review and control how images are manipulated to avoid misinformation. Further research on detecting machine-generated images is needed to mitigate this potential issue.

\section{Conclusion}
In this paper, we propose IIR-Net, a text-to-image editing model that incorporates the original image by selectively erasing the image information. IIR-Net mainly consists of two stages: an conditional diffusion model that takes the original image as additional control, and an image information removal module to address the identical mapping issue. We demonstrate that IIR-Net outperforms the state-of-the-art in both qualitative and quantitative evaluations on CUB, Outdoor Scenes, and COCO datasets. For example, compared to Imagic, IIR-Net improves the LPIPS score from 0.57 to 0.30 and the CLIP score from 21.53 to 24.30 on COCO, with a speed improvement of two orders of magnitude. We also use qualitative examples to demonstrate the effectiveness of our model on various image editing tasks, validating that our model can modify the target attribute according to language descriptions while preserving the text-irrelevant content of the original image well.

\noindent\textbf{Acknowledgements}
This material is based upon work supported, in part, by DARPA under agreement number HR00112020054. Any opinions, findings, and conclusions or recommendations are those of the author(s) and do not necessarily reflect the views of the supporting agencies.

%%%%%%%%% REFERENCES
{\small
\bibliographystyle{ieee_fullname}
\bibliography{egbib}
}

\appendix

\section{Additional Experimental Results}
\label{sec:additional_results}
\begin{figure*}[t]
    \centering
    \includegraphics[width=.85\linewidth]{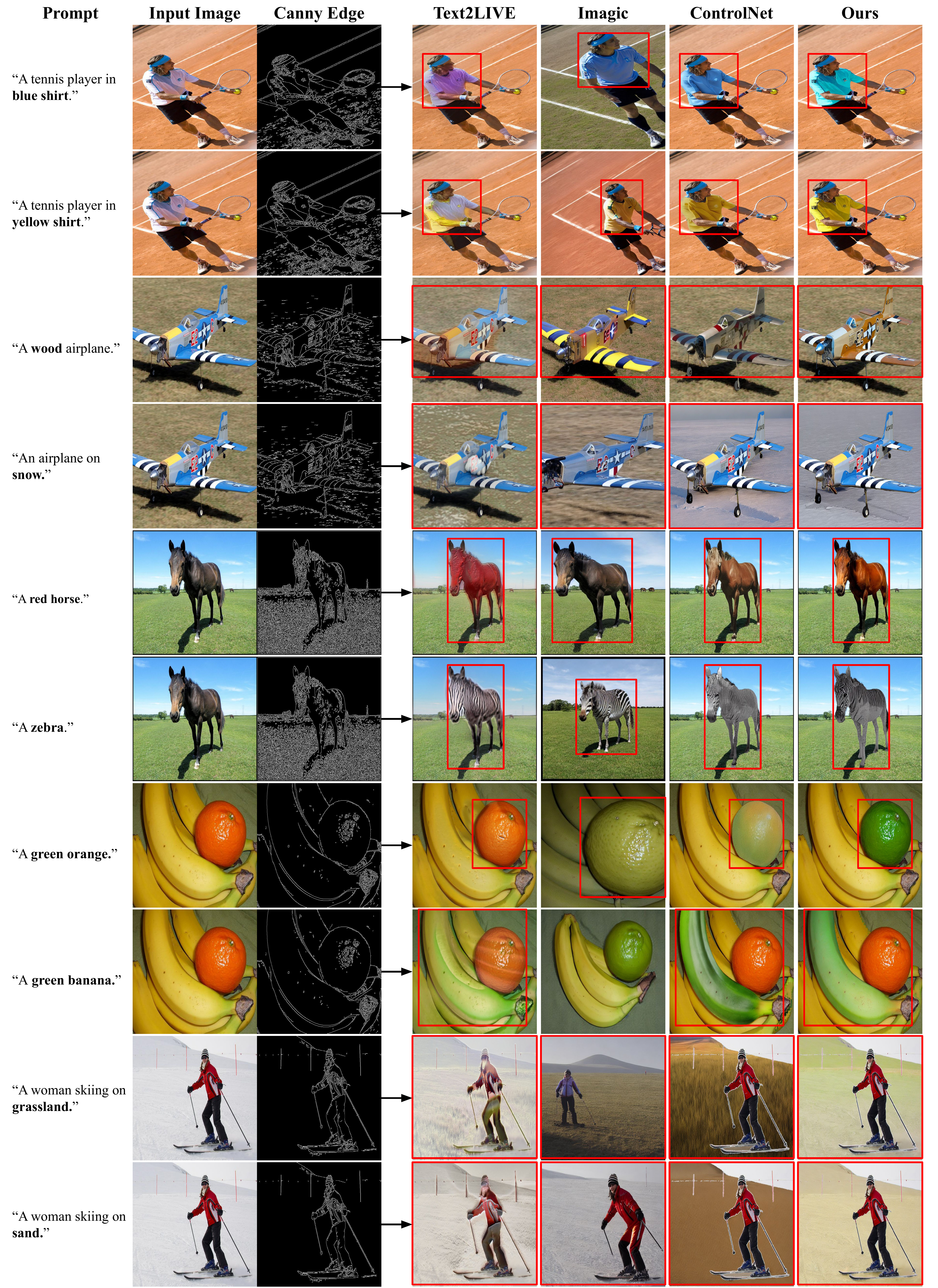}
    \caption{Additional comparison results between IIR-Net and the baselines on the COCO dataset. We set the image resolution to 512$\times$512. We observe that our method effectively modifies the input image while preserving the text-irrelevant content. For instance, in the example of ``A tennis player in \textbf{blue shirt},'' IIR-Net retains both the shape and texture attributes of the original shirt, whereas the other baselines either introduce limited visual effects or modify text-irrelevant content such as textures or shape. See \cref{sec:additional_results} for further discussion.} 
    \label{fig:supplementary_figure1}
\end{figure*}

\begin{figure*}[t]
    \centering
    \includegraphics[width=.85\linewidth]{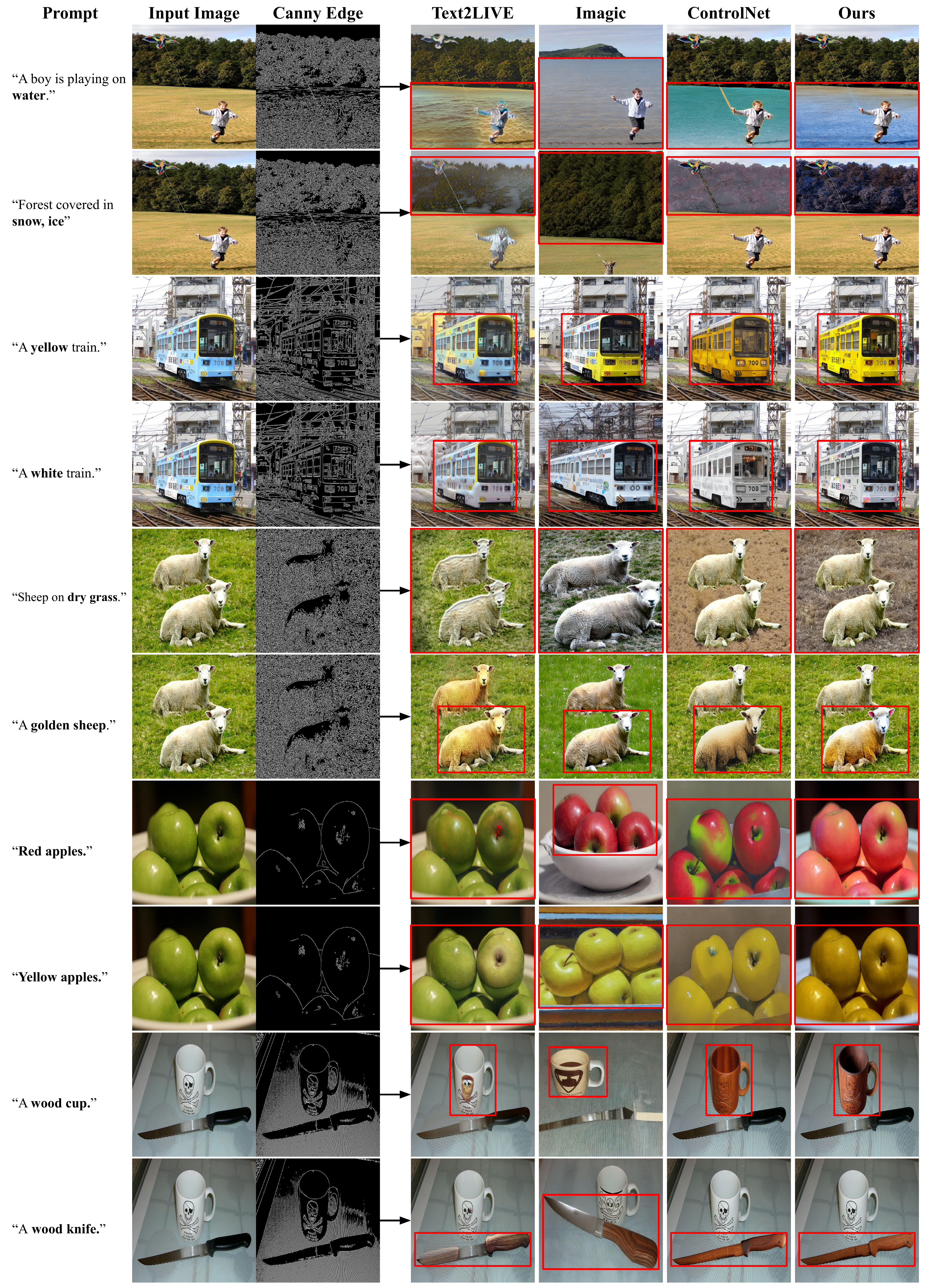}
    \caption{Additional comparison results between IIR-Net and baselines on the COCO dataset. We set the image resolution to 512$\times$512. See \cref{sec:additional_results} for discussion.} 
    \label{fig:supplementary_figure2}
\end{figure*}

We present additional qualitative results in \Cref{fig:supplementary_figure1} and \Cref{fig:supplementary_figure2} to supplement the main paper. The results demonstrate that IIR-Net can modify image content base on user prompts while preserving the text-irrelevant content of the original image. In \Cref{fig:supplementary_figure1}, we observe that IIR-Net successfully preserves shape-related information of the target object in texture editing examples (\eg, ``A \textbf{wood} airplane'' and ``A woman skiing on \textbf{grassland}''), as well as texture-related information in color editing examples (\eg, ``A \textbf{red} horse.'' and ``A \textbf{green} orange''). In contrast, Imagic~\cite{kawar2023imagic} may modify the shape information, while ControlNet~\cite{zhang2023adding} may modify the texture information. Besides, we observe that our network produces visually more natural images compared to Text2LIVE~\cite{bar2022text2live}. \Eg, in the example of ``A \textbf{red} horse,'' Text2LIVE applies some red effects to the horse, whereas our method directly produces ``a red horse'' with better consistency to the background. These observations are consistent with our conclusions in the main paper.

\section{User study Interface}
\label{sec:user_study_template}
\begin{figure*}[t]
    \centering
    \includegraphics[width=.8\linewidth]{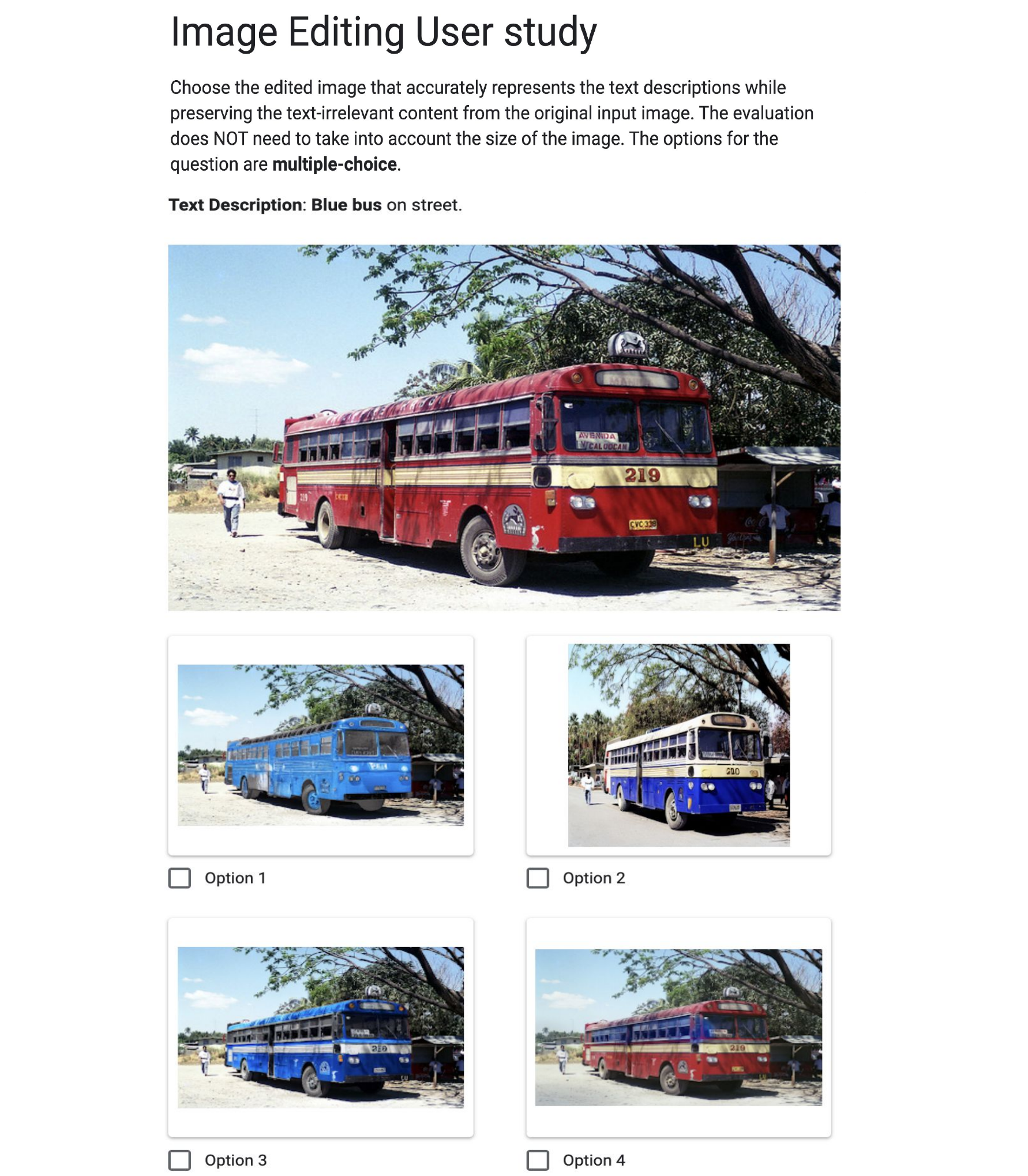}
     \caption{\textbf{User study screenshot.} A sample screenshot illustrating one of the questions presented to participants in our user study. See \cref{sec:user_study_template} for discussion.} 
    \label{fig:supplementary_user_study_template}
\end{figure*}

In our user study experiments, annotators were presented with an input image, a target text, and four edited images generated by different methods. They were asked to evaluate the accuracy of manipulated images according to two aspects: (1) the alignment of the image with the target text, and (2) the preservation of text-irrelevant content from original images. We provide a sample screenshoot in \Cref{fig:supplementary_user_study_template}.

%-------------------------------------------------------------------------

% % \clearpage
% %%%%%%%%% REFERENCES
% {\small
% \bibliographystyle{ieee_fullname}
% \bibliography{egbib}
% }

\clearpage

\end{document}